\documentclass[conference]{IEEEtran}
\IEEEoverridecommandlockouts
\usepackage{amsmath,amssymb,amsfonts}
\usepackage{algorithmic}
\usepackage{graphicx}
\usepackage{textcomp}
\usepackage{xcolor}
\usepackage{orcidlink}
\usepackage{comment}
\usepackage[backend=bibtex,style=numeric,sorting=none,maxnames=5,minnames=5]{biblatex}
\def\BibTeX{{\rm B\kern-.05em{\sc i\kern-.025em b}\kern-.08em T\kern-.1667em\lower.7ex\hbox{E}\kern-.125emX}}
    
\begin{document}

\title{A Multimodal Conversational Assistant for the Characterization of Agricultural Plots from Geospatial Open Data
\thanks{This work was partially supported by Project PID2021-124383OB-I00 of the Spanish National Plan for Research, Development and Innovation.}
}

\author{\IEEEauthorblockN{Juan Cañada*\orcidlink{0009-0009-7223-9815}}
\IEEEauthorblockA{\textit{Dept. AI $\&$ Data Analytics} \\
\textit{CTIC Technology Center}\\
Gijón, Spain \\
juan.canada@fundacionctic.org
}
\and
\IEEEauthorblockN{Raúl Alonso}
\IEEEauthorblockA{\textit{Dept. IT} \\
\textit{CTIC Technology Center}\\
Gijón, Spain \\
raul.alonso@fundacionctic.org}
\and
\IEEEauthorblockN{Julio Molleda \orcidlink{0000-0002-2949-1195}}
\IEEEauthorblockA{\textit{Dept. Comp. Sci. and Eng.} \\
\textit{University of Oviedo}\\
Gijón, Spain \\
jmolleda@uniovi.es}
\and
\IEEEauthorblockN{Fidel Díez \orcidlink{0000-0003-4523-1577}}
\IEEEauthorblockA{{\textit{Dept. R$\&$D}} \\
\textit{CTIC Technology Center}\\
Gijón, Spain \\
fidel.diez@fundacionctic.org}
}

\maketitle

\begin{abstract}
The increasing availability of open Earth Observation (EO) and agricultural datasets holds great potential for supporting sustainable land management. However, their high technical entry barrier limits accessibility for non-expert users. This study presents an open-source conversational assistant that integrates multimodal retrieval and large language models (LLMs) to enable natural language interaction with heterogeneous agricultural and geospatial data. The proposed architecture combines orthophotos, Sentinel-2 vegetation indices, and user-provided documents through retrieval-augmented generation (RAG), allowing the system to flexibly determine whether to rely on multimodal evidence, textual knowledge, or both in formulating an answer. To assess response quality, we adopt an LLM-as-a-judge methodology using Qwen3-32B in a zero-shot, unsupervised setting, applying direct scoring in a multi-dimensional quantitative evaluation framework. Preliminary results show that the system is capable of generating clear, relevant, and context-aware responses to agricultural queries, while remaining reproducible and scalable across geographic regions. The primary contributions of this work include an architecture for fusing multimodal EO and textual knowledge sources, a demonstration of lowering the barrier to access specialized agricultural information through natural language interaction, and an open and reproducible design. 
\end{abstract}


\section{Introduction and related works}\label{sec/1}

Earth Observation (EO) programs have generated a vast volume of open-source geospatial data in recent years. Initiatives such as the Copernicus program with Sentinel satellites or national mapping agencies make available extensive repositories of optical, multispectral, and thematic data. At the same time, an increasing number of open-access documents, reports, and scientific publications related to agriculture and land management are freely accessible. However, despite their availability, these resources remain underutilized due to the high entry barrier: accessing, processing, and interpreting EO data requires specialized knowledge, technical skills, and dedicated software pipelines. Consequently, non-expert users—such as farmers, local stakeholders, or small companies—often lack the ability to extract actionable insights from this information.

With the rapid development of artificial intelligence, and in particular the emergence of multimodal large language models (MLLMs), agricultural practices have encountered a major opportunity for transformation. Several approaches have attempted to use these technologies to bridge the data accessibility gap by developing conversational or multimodal assistants in specific agricultural contexts. The authors of \cite{koopman_agask_2024} and \cite{silva_gpt-4_2023} illustrate how large language models (LLMs) can answer farmers’ questions from scientific or agronomic documents, while \cite{zhao_chatagri_2023} explores ChatGPT-based interactions for agricultural queries. In the visual domain, \cite{wang_agri-llava_2024} extends multimodal LLMs to crop pest and disease identification. Beyond agriculture, the potential of dialog systems grounded in EO data has been shown in works such as \cite{soni_earthdial_2025}, \cite{xu_multimodal_2025}, and \cite{kuckreja_geochat_2023}, which explore multisensory EO-to-dialog pipelines. Although these systems demonstrate the promise of conversational interfaces in their respective domains, they are often limited to a single modality or a narrow scope, and do not fully exploit the breadth of heterogeneous open EO data available.

\begin{figure}[t]
\centerline{\includegraphics[width=\columnwidth]{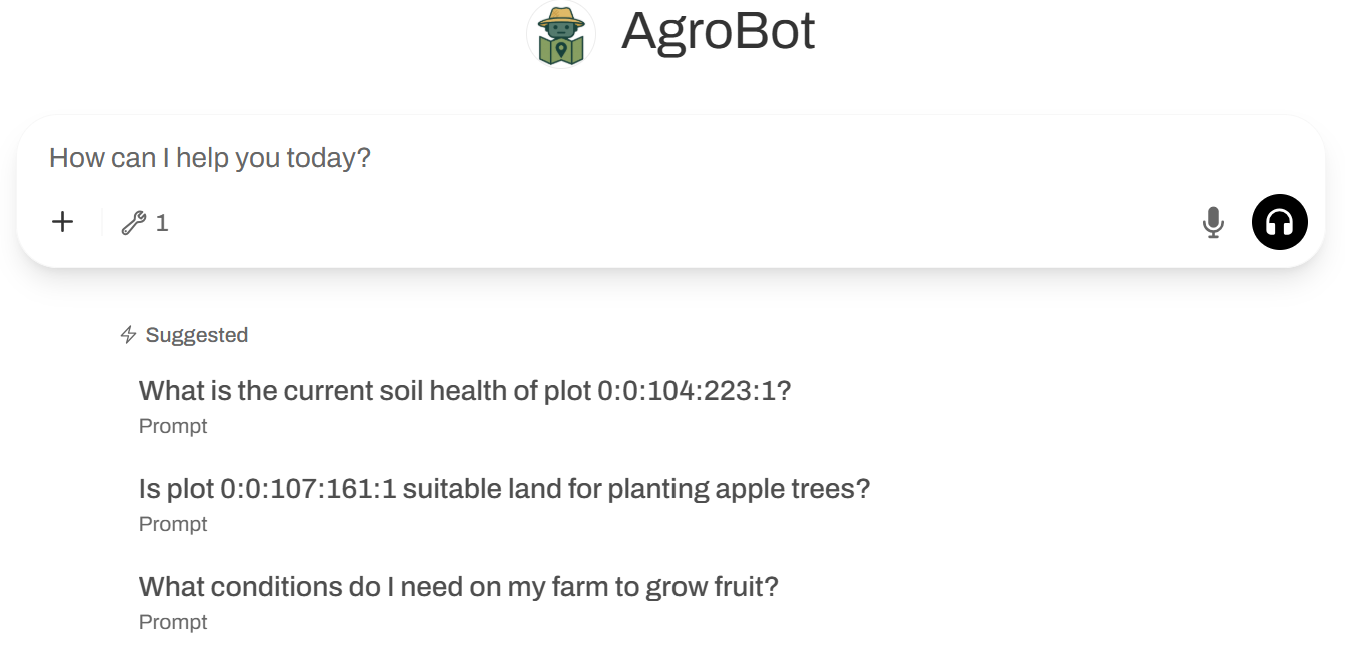}}
\caption{Conversational assistant user interface (developed over OpenWebUI \cite{owui}). It includes input text box, voice input button, and suggested prompts.}
\label{fig:interface}
\end{figure}

\begin{figure}[t]
\centerline{\includegraphics[width=\columnwidth]{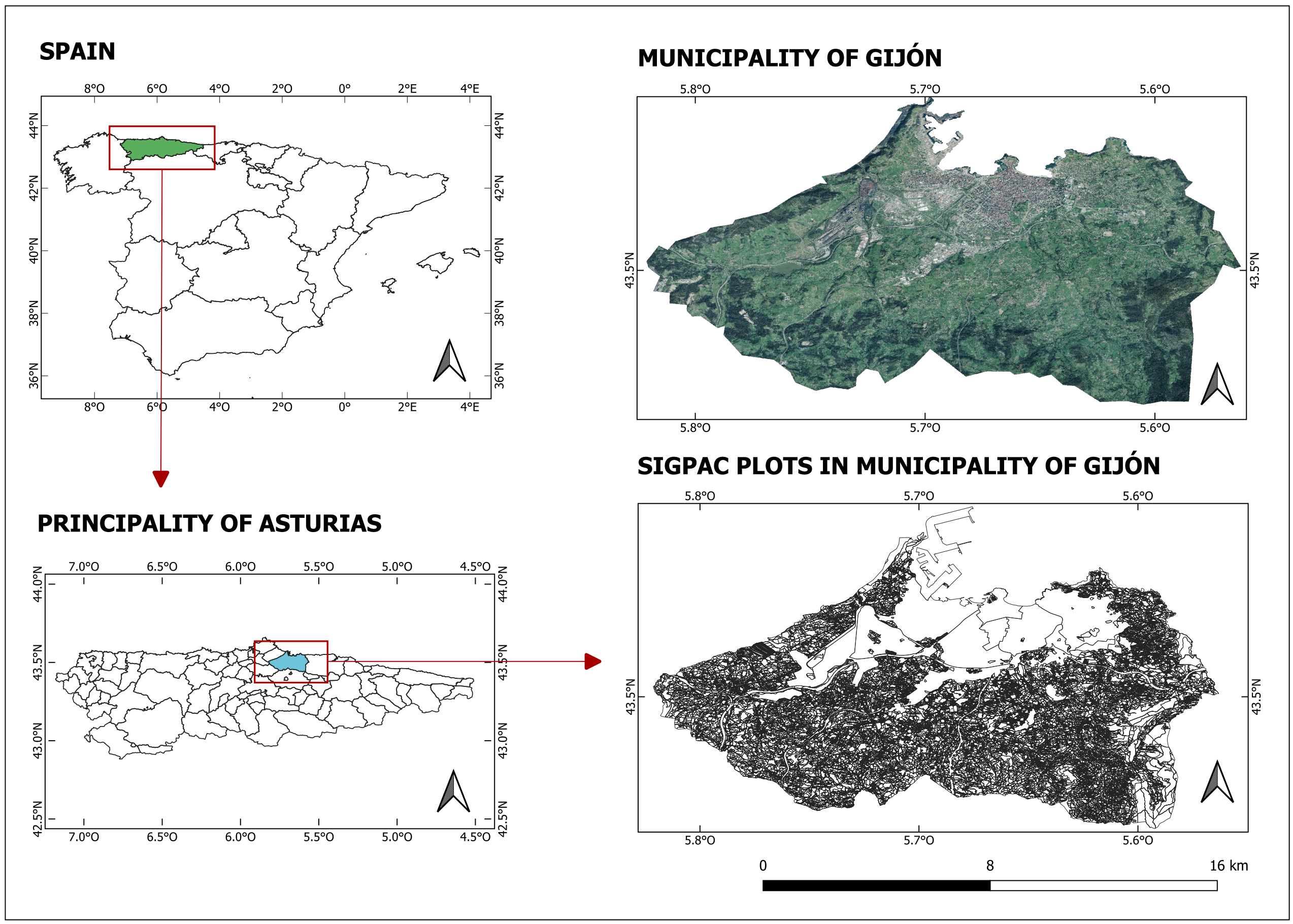}}
\caption{The study region is the municipality of Gijón, in Asturias, Spain, with an area of approximately 182 km$^{2}$. SIGPAC registry is the official system for agriculture and farmer parcel identification.}
\label{fig:map}
\end{figure}

In this work, we present a conversational assistant (Fig. \ref{fig:interface}) that integrates heterogeneous open data sources for agriculture and territorial analysis. The system combines (i) descriptive information of the terrain given by  multimodal LLM processing of 25 cm-resoution ortophotos, (ii) multispectral Sentinel-2 imagery to derive vegetation indices such as NDVI or EVI, and (iii) retrieval-augmented generation (RAG) over domain-specific documents, including both public repositories and user-provided PDFs. By aggregating these diverse data streams, the assistant allows users to query a parcel of land in natural language and receive a contextualized, multi-source answer.
Our contribution lies in the architecture and integration strategy of this system: we demonstrate how multimodal LLMs, open EO pipelines, and retrieval methods can be orchestrated into a single interface that lowers the barrier to access and creates value from open geospatial data. The approach is validated through preliminary use cases, showing its potential to support decision-making in agriculture and land management. The study area for this pilot is the municipality of Gijón, in northern Spain (Fig. \ref{fig:map}), which has a mixed urban and agricultural landscape. The delimitation of the agricultural plots is given by the SIGPAC (Geographic Information System for Agricultural Plots) registry, created by the spanish Ministry of Agriculture, Fisheries and Food.

\section{System Architecture} \label{sec/2}

\begin{figure*}[t!]
\centering{\includegraphics[width=\textwidth]{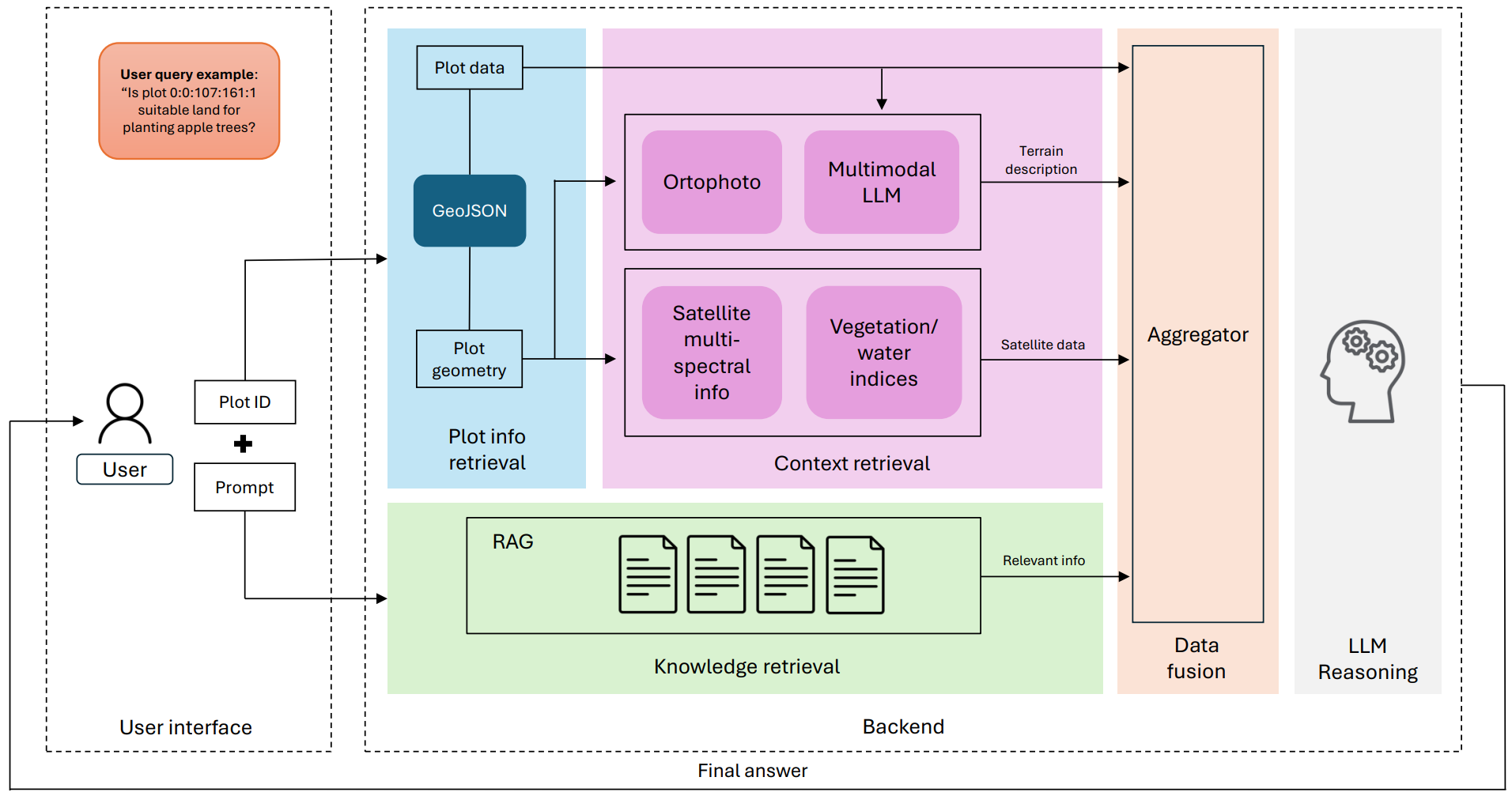}}
\caption{The system architecture is composed of two main modules: the conversational interface, where the user-assistant interaction takes place, and the backend processing of the query. Several blocks, responsible for multimodal data retrieving, are interconnected, and then fused through a multi-domain aggregator, which delivers the processed context to the final LLM.}
\label{fig:arch}
\end{figure*}

The proposed system is designed to combine several interconnected modules, which integrate heterogeneous data sources. The overall architecture, included in Fig. \ref{fig:arch}, is composed of two main layers: user interface and backend processing, which includes the LLM reasoning. The  aim of the data flow is to provide the final LLM with the most rich context regarding the area of interest (a specific parcel) in order for it to give the best possible answer to the user's query. 

\subsection{User interface}
The interaction with the system takes place through a conversational interface, where the user submits a query in natural language along with a plot identifier (Plot ID). This ID is obtained by concatenating the Province, Municipality, District, Zone, Polygon, and Parcel field codes in SIGPAC. For example: “Is plot 0:0:107:161:1 adequate land for planting apple trees?”. The interface includes voice-to-text mode, contributing to its accessibility.

\subsection{Backend}

The backend consists of five different blocks, which conform a data stream from the user query to the final LLM, and then back to the user with the answer: (i) plot info retrieval, which is responsible for fetching, from the SIGPAC registry, both plot data -as slope, elevation, area, perimeter or land use-, and the geo-referenced geometry. Once known the area of interest, the (ii) context retrieval block performs two parallel tasks, each leveraging different open-source data sources to enrich the context: (a) an ortophoto of the plot, obtained from aerial imagery, is processed through a multimodal LLM along with the plot data, generating a detailed description of the terrain. In addition, (b) satellite multispectral data, specifically from Sentinel-2 (Copernicus mission, European Spacial Agency) are retrieved. After spacial and temporal processing combining multiple spectral bands, an array of vegetation and water indices that reflect biophysical conditions of the land are obtained. In parallel, (iii) the system queries a document repository through a retrieval-augmented generation mechanism, which provides relevant textual knowledge such as agronomic guidelines, regulations, or best practices. All these multi-domain data are then fused through (iv) an aggregator, which converts all data to natural language, and prepares the information into an encapsulated prompt that provides the (v) final LLM with the overall context of the plot and the query. Several open-source models were evaluated, including Qwen3-32B, Llama-3.3-70B or Mistral-Small-3.2-24B, being Qwen the one chosen for the final implementation due to its superior performance. Regarding the LLMs framework, the model inference is managed with the inference engine vLLM \cite{vllm}, orchestrated through the liteLLM proxy \cite{litellm}, which agglutinates all models in one port. Besides from the multimodal and final LLMs, there are also embeddings, reranker, and tool-triage models working together. The infrasctructure runs in a 8x80GB cluster of NVIDIA GPUs.

\section{Assessment} \label{sec/3}

\begin{figure}[t]
\centerline{\includegraphics[width=\columnwidth]{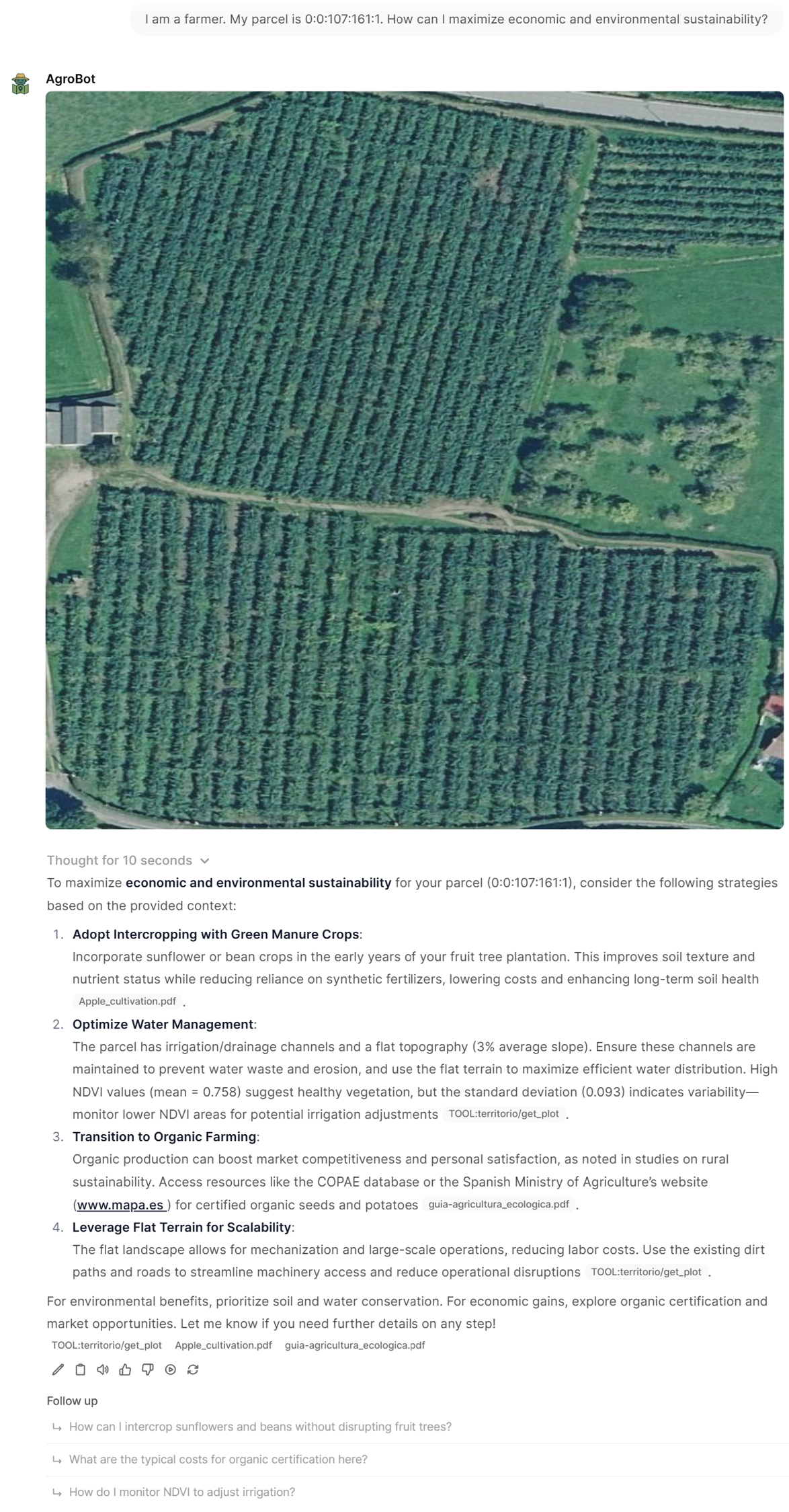}}
\caption{Example of an assistant response to user's query"I am a farmer. My parcel is 0:0:107:161:1. How can I maximize economic and environmental sustainability?".}
\label{fig:response}
\end{figure}

Evaluating the quality of the response in domain-specific scenarios, as geospacial and agriculture, is challenging, as no significative benchmarks are available. In this work, we adopt an LLM-as-a-judge methodology \cite{llm-as-a-judge}, which has become a standard practice for preliminary assessment of LLM responses. After testing different variants of the Qwen model family, such as Qwen3-8B-AWQ, which applies activation-aware weight quantization, or Qwen3-4B, we decided to employ the Qwen3-32B model, which delivers better-quality answers compared to its smaller versions. The evaluation was performed in an unsupervised, zero-shot setup, where the judge model was provided with a prompt explaining the assessment criteria but no example answers. Following an approach of direct scoring in multi-dimensional quantitative evaluation, the LLM-as-a-judge was fed with the user's query, the retrieved context and the assistant's answer. Each system response was assessed through four dimensions: \textit{correctness} (accuracy with respect to the context), \textit{relevance} (focus on the query), \textit{clarity} (readability and accessibility), and \textit{completeness} (extent to which the user's query is fully answered). The judge provided scores on a 1-5 scale, along with a short justification. 

\subsection{Preliminary results}

The evaluation process included 45 experiments. In order to assess separately both retrieval techniques, 30 experiments focused on triggering multimodal retrieval and 15 were RAG-based queries\footnote{A tool-triage model is responsible for deciding whether to use one, another, both or neither of the them, being the multimodal retrieval only applicable when providing a plot ID.}. Table \ref{tab:stats-results} contains the average score of the four evaluation dimensions for the 45 experiments, grouped into multimodal and RAG, and the global average. For the seek of simplicity, five queries, along with their evaluation, are presented in Table \ref{tab:5results}. The first of them is RAG-based, while the remaining four rely on multimodal retrieval. Table \ref{tab:2results} in the appendix complements these results, extending a comprehensive evaluation for two of the experiments considered relevant (one for each query focus mode) with all the information: user's query, retrieved context, assistant's response and LLM-as-a-judge assessment. Regarding the interface display, Fig. \ref{fig:response} includes a system response to the query "I am a farmer. My parcel is 0:0:107:161:1. How can I maximize economic and environmental sustainability?". The response layout is divided into the following elements: the parcel ortophoto is fetched and embedded in the answer as a base64 image, the response is formatted in Markdown and displayed below, along with the citations of the retrieved context, and finally a "Follow up" section with suggested prompts is included at the bottom. 

\begin{table*}[t]
\caption{Average score in multi-dimensional quantitative evaluation for the 45 experiments, grouped into Multimodal retrieval (30) and RAG-based (15).}
\begin{center}
\begin{tabular}{p{0.47\textwidth}p{0.1\textwidth}p{0.1\textwidth}p{0.1\textwidth}p{0.1\textwidth}}
\hline
\textbf{Query Focus Mode} & \textbf{Correctness} & \textbf{Relevance} & \textbf{Clarity} &\textbf{Completeness}\\
\hline
Multimodal retrieval & 4.867 & 4.767 & 4.567 & 4.367 \\
RAG-based & 4.733 & 4.400 & 4.267 & 4.333 \\
Total & 4.800 & 4.583 & 4.417 & 4.350 \\
\hline
\end{tabular}
\label{tab:stats-results}
\end{center}
\end{table*}


\begin{table*}[t]
\caption{Query and assessment results for five examples. The first one is RAG-based, and the next four are focused on multimodal retrieval.}
\begin{center}
\begin{tabular}{p{0.2\textwidth}p{0.75\textwidth}}
\hline
\textbf{User's query} & \textbf{Assessment} \\
\hline
"What conditions do I need on my farm to grow fruit?"
&
"correctness: 5, relevance: 5, clarity: 4, completeness: 5, justification: ""The response accurately extracts key conditions for fruit cultivation from the context (e.g., cultivar selection, pollination, rootstocks, NAA application) and addresses the user’s query. While most technical terms are explained, concepts like 'clonal rootstocks' or 'renewal pruning' could be clarified further for non-experts. It integrates all critical context elements, including challenges like spring frost and resource inefficiency.""\\
"Is plot 0:0:107:55:1 suitable land for planting apple trees?"
&
"correctness: 5, relevance: 5, clarity: 5, completeness: 5, justification: The response correctly identifies the plot as pastureland with a 21.6$\%$ slope, aligns with the NDVI data to highlight vegetation health, and provides actionable recommendations (rootstocks, irrigation, terracing) while addressing the user's question about apple tree suitability and integrating all key context elements.""\\
"What is the current soil health of plot 0:0:104:223:1?"
&
"correctness: 5, relevance: 5, clarity: 5, completeness: 5, justification: The response accurately acknowledges the absence of direct soil health metrics (pH, nutrients) in the context while using NDVI data (mean 0.868, max 0.903) and terrain features (drainage channels, water bodies) to infer indirect indicators of soil health. It clearly explains limitations and connects vegetation health to pastureland management, ensuring both relevance and completeness.""\\
"I want to buy terrain for planting raspberries. Is plot 0:0:105:9000:1 a good option?"
&
"correctness: 4, relevance: 5, clarity: 4, completeness: 5, justification: "The response correctly identifies the urban classification, NDVI values, and potential infrastructure challenges but assumes soil type (loamy/organic matter) not explicitly stated in the context. It addresses raspberry-specific concerns like zoning and climate compatibility while integrating key context elements (NDVI, slope, altitude).""\\
"I am a farmer. My parcel is 0:0:107:161:1. How can I maximize economic and environmental sustainability?"
&
"correctness: 4, relevance: 5, clarity: 4, completeness: 5, justification: The response correctly uses the flat terrain (3$\%$ slope), irrigation channels, and NDVI metrics (mean 0.758, stdDev 0.093) to suggest intercropping, water optimization, and mechanization. It links environmental sustainability to soil/water conservation and economic gains to organic certification. However, assumptions like intercropping with sunflower/beans and specific resources (COPAE, MAPA) are not explicitly supported by the context, slightly reducing correctness. Clarity is strong but could simplify terms like 'green manure.""\\
\hline
\end{tabular}
\label{tab:5results}
\end{center}
\end{table*}

\section{Discussion} \label{sec/4}

The results presented in Table \ref{tab:stats-results} show that overall, the system produces top-quality responses. Dimensions \textit{correctness} and \textit{relevance} are the ones with the higher average score, achieving 4.5+ out of 5 points, while \textit{clarity} and \textit{completeness} remain slightly lower. Regarding the query focus mode, multimodal retrieval-based queries tend to achieve better quality responses than RAG-based queries across all evaluation dimensions, being \textit{relevance} the largest difference. Table \ref{tab:5results} depicts five different user queries that produce a high quality answer. It can be seen how the system is able to respond consistently to different topics and user intentions, as planting trees, maximizing environmental sustainability or evaluating a parcel for its potential purchase. Given the specificity of the action domain of the system, the degree of completeness and correctness of the answers inherently relies on how well the query is aligned with the assistant's capabilities.  For instance, a request about local fauna would yield a less complete response compared to an agriculture-based query. Likewise, RAG-based queries depend on the provided knowledge database, which we consider a strength, enabling flexible customization for every user need. Table \ref{tab:2results} allows to see how the model is able to retrieve relevant context regarding the user's query, for both modalities, and produce an accurate response. The multimodal retrieval example exhibits the capacity of the model to fetch a generated-description of the terrain, and vegetation indices information, to enrich the quality of the answer.

\section{Conclusion} \label{sec/5}
This study aimed to explore how open-source LLMs and multimodal retrieval can be combined into a conversational assistant for agricultural and geospatial applications. By lowering the barrier of access to complex EO and agronomic data, the proposed system enables users to query heterogeneous open data sources through natural language interactions. The primary contributions of this work include: (i) an architecture that fuses information from multiple modalities and domains —satellite imagery, vegetation indices, orthophotos, and textual documentation— into a unified conversational framework, (ii) a demonstration of how open-source LLMs can make highly specialized EO and agricultural information more accessible, delivering precise and context-aware answers for land management tasks, and (iii) an open, reproducible and scalable design. The presented results demonstrate that our proposed system is capable of producing complete, accurate, clear and correct responses to agro-related queries by retrieving and integrating relevant information regarding each desired plot, with LLM-as-a-judge evaluation confirming the utility of integrating multimodal and RAG-based retrieval strategies. Although this study was conducted in a specific region, the design is not tailored to local landscape features, making it both reproducible in other geographic areas and scalable to larger regions. Although the evaluation remains preliminary, it provides evidence that conversational interfaces can effectively bridge the gap between highly technical EO datasets and end users without specialized training. Future work will focus on expanding to more context retrieval sources, optimizing the computational efficiency, and assessing further state-of-the-art open-source models.

\section*{CRediT author statement}

CRediT roles and contributions: \textbf{Juan Cañada:} Conceptualization, Data curation, Formal Analysis,  Investigation, Methodology, Software, Validation, Visualization, and Writing – original draft. \textbf{Raúl Alonso:} Resources and Writing – review $\&$ editing. \textbf{Julio Molleda: } Writing – review $\&$ editing and Supervision. \textbf{Fidel Díez:} Funding acquisition, Project administration and Writing – review $\&$ editing.


\printbibliography

\appendix

\section{Appendix}

\begin{table*}[t]
\caption{Assessment results are divided into two groups: Multimodal retrieval cases, and RAG-based cases (although it is common that both are included at the same time). An example of each case is listed in the table.}
\begin{center}
\begin{tabular}{p{0.46\textwidth} p{0.46\textwidth}}
\hline
Multi-modal retrieval example & RAG example  \\
\hline

\textbf{User's query}

Is plot 0:0:107:55:1 suitable land for planting apple trees? 

\textbf{Retrieved context}

Terrain description: The image shows a plot of land classified as "PASTIZAL" (pastureland), which typically indicates grassy areas used for grazing livestock. Here are the geographic and terrain features visible in the image:

1. Grassy Field: The majority of the plot is covered with green grass, which is characteristic of pastureland. The grass appears to be well-maintained and uniform in color, suggesting it is actively managed.
2. Dirt Paths or Roads: There are several winding dirt paths or roads that traverse the plot. These paths are likely used for accessing different parts of the pasture, moving livestock, or for maintenance purposes.
3. Vegetation Patches: There are a few patches of denser vegetation or possibly small trees scattered throughout the plot. These could be natural features or planted for shade or windbreaks.
4. Topography: The terrain appears relatively flat with slight undulations. The paths follow the natural contours of the land, which suggests gentle slopes rather than steep terrain.
5. Water Bodies or Wet Areas: There are a few darker areas that could indicate wetter sections of the pasture, possibly seasonal water accumulation or low-lying areas that retain moisture.
6. Fencing: There is no clear evidence of fencing in the image, but the paths might be used to guide livestock within the pasture.

Overall, the image depicts a typical pastureland with grassy areas, paths for access, and some patches of denser vegetation, set on relatively flat terrain..

Area in ha = 0.763,
Perimeter in meters = 375.35,
Average slope = 21.6 $\%$,
Altitud in meters= 94,
Land use (in spanish) = PASTIZAL.
The NDVI statistics are 'NDVI$\_$max': 0.9097, 'NDVI$\_$mean': 0.8468, 'NDVI$\_$min': 0.6066, 'NDVI$\_$stdDev': 0.0527) in range [0-1].

\textbf{Response}
The plot 0:0:107:55:1 is classified as pastureland ("PASTIZAL") with a 21.6$\%$ average slope, which may pose challenges for apple cultivation. Apple trees typically require flat to gently sloping terrain (ideally <15-20$\%$) for proper irrigation and erosion control.

However, the high NDVI values (mean 0.847, max 0.91) indicate healthy vegetation and fertile soil, which are beneficial for apple growth 

To determine suitability, consider:

- Rootstock selection: Rootstocks like MM 111 or Merton 793 are recommended for sloped or hilly areas and offer drought tolerance and pest resistance.
- Water management: The plot would need a robust irrigation system (e.g., drip irrigation) to address the higher slope and ensure consistent water supply.
- Soil preparation: Terracing or contour farming might be necessary to prevent erosion and optimize water retention.
While the land has favorable vegetation health, the slope and lack of existing fencing suggest careful planning would be required for successful apple cultivation.

\textbf{Assessment}
 
"correctness": 5,
"relevance": 5,
"clarity": 5,
"completeness": 5,
"justification": "The response correctly identifies the plot as pastureland with a 21.6$\%$ slope, aligns with the NDVI data to highlight vegetation health, and provides actionable recommendations (rootstocks, irrigation, terracing) while addressing the user's question about apple tree suitability and integrating all key context elements."

&

\textbf{User's query}

What conditions do I need on my farm to grow fruit?

\textbf{Retrieved context}

"Relevance
82.21$\%$ (0.8221)
Content
Life cycle of apple fruit crop
Source
Apple$\_$cultivation.pdf (page 68)
Relevance
79.67$\%$ (0.7967)
Content
Constraint in apple cultivation
Large number of old orchards (more than 30 years old) are showing decline in terms of growth and fruit yield. 
Choice of wrong pollinizers and their inadequacy in number often result to low productivity .
The need for injecting new blood into the apple industry through spread of new cultivars (spur types, color mutants, Red Fuji; scab resistant cultivars, bud sport selections of Royal Delicious, and some of the promising hybrids) is urgently felt .
The low chilling cultivars and promising cultivars identified need further spread. 
Technologies like use of clonal rootstocks, introduction of renewal pruning techniques and micro nutrient applications have not been transferred and adopted at a satisfactory level.
The water and fertilizer use efficiency is generally poor . Also, spring frost and hailstorms are adverse weather parameters leading to low productivity
Source
Apple$\_$cultivation.pdfº (page 51)
Relevance
77.86$\%$
(0.7786) Content
Plant growth regulator
Use of growth hormones is essential for good flowering and proper colouration in fruits.
NAA @ 10 -20 ppm at petal fall may be applied for the purpose."

\textbf{Response}
"To grow fruit successfully on your farm, consider the following conditions based on the context:

Cultivar Selection: Use modern, high-yielding cultivars (e.g., Red Fuji, scab-resistant varieties) and avoid old orchards beyond 30 years, which show declining productivity.
Pollination: Ensure suitable pollinizers are planted in adequate numbers to avoid low productivity due to poor pollination.
Climate Adaptation: Address adverse weather risks like spring frost and hailstorms, which reduce yields. Low-chill cultivars may be suitable if your region has mild winters.
Agronomic Practices: Adopt technologies such as clonal rootstocks, renewal pruning, and micronutrient applications to improve growth and fruit quality.
Resource Efficiency: Optimize water and fertilizer use, as inefficiencies in these areas are common challenges in fruit farming 
Growth Regulators: Apply plant growth regulators like NAA (10–20 ppm) at petal fall to enhance flowering and fruit coloration [2].

Adjust these factors to your local conditions and specific fruit type. Let me know if you need details on a particular aspect!
"

\textbf{Assessment}
 
"correctness": 5,"relevance": 5, "clarity": 4, "completeness": 5,
"justification": "The response accurately extracts key conditions for fruit cultivation from the context (e.g., cultivar selection, pollination, rootstocks, NAA application) and addresses the user’s query. While most technical terms are explained, concepts like 'clonal rootstocks' or 'renewal pruning' could be clarified further for non-experts. It integrates all critical context elements, including challenges like spring frost and resource inefficiency."
\\

\hline
\end{tabular}
\label{tab:2results}
\end{center}
\end{table*}


\end{document}